%% file: main.tex
\pdfoutput=1

\documentclass[twocolumn]{IEEEtran}

\usepackage[utf8]{inputenc} %
\usepackage[T1]{fontenc}    %
\usepackage[colorlinks]{hyperref}       %
\usepackage{url}            %
\usepackage{booktabs}       %
\usepackage{amsfonts}       %
\usepackage{nicefrac}       %
\usepackage{microtype}      %
\usepackage{amsmath}
\DeclareMathOperator{\sigmoid}{\sigma}
\usepackage{graphicx}
\usepackage[autostyle]{csquotes}
\allowdisplaybreaks
\usepackage{caption}

\newcommand{\uproman}[1]{\uppercase\expandafter{\romannumeral#1}}

\usepackage[normalem]{ulem}
\usepackage{xcolor}

\title{Engineering the Neural Automatic Passenger Counter}

\author{
\IEEEauthorblockN{Nico Jahn\IEEEauthorrefmark{1}, Michael Siebert\IEEEauthorrefmark{1}\\
\IEEEauthorblockA{Interautomation Deutschland GmbH, \\
Berlin, Germany\\
\{nico.jahn, michael.siebert\}@interautomation.de\\
\IEEEauthorrefmark{1} All authors contributed equally to this work}}
}

\usepackage[
	style=ieee,
	backend=biber,
	url=true,
	isbn=false,
	doi=false,
	date=year,
	maxnames=6,
	mincitenames=2,
	maxcitenames=2, %
]{biblatex}
\begin{document}
\captionsetup[figure]{name={Figure}}

\maketitle

\begin{abstract}
Automatic passenger counting (APC) in public transportation has been approached with various machine learning and artificial intelligence methods since its introduction in the 1970s. While equivalence testing is becoming more popular than difference detection (Student's t-test), the former is much more difficult to pass to ensure low user risk. On the other hand, recent developments in artificial intelligence have led to algorithms that promise much higher counting quality (lower bias). However, gradient-based methods (including Deep Learning) have one limitation: they typically run into local optima. In this work, we explore and exploit various aspects of machine learning to increase reliability, performance, and counting quality. We perform a grid search with several fundamental parameters: the selection and size of the training set, which is similar to cross-validation, and the initial network weights and randomness during the training process. Using this experiment, we show how aggregation techniques such as ensemble quantiles can reduce bias, and we give an idea of the overall spread of the results. We utilize the test success chance, a simulative metric based on the empirical distribution. We also employ a post-training Monte Carlo quantization approach and introduce cumulative summation to turn counting into a stationary method and allow unbounded counts.
\end{abstract}


\input{chapters/introduction}
\input{chapters/methods}
\input{chapters/results}
\input{chapters/conclusion}

\section{Acknowledgements}
This research is financially supported by the European Regional Development Fund under Grant 10166961.

\printbibliography

\end{document}

%% file: chapters/introduction.tex
\section{Introduction}
Assessment of passenger counts is of paramount importance for public transport agencies to plan, manage and evaluate their transit service. Over the past three decades, automatic passenger counting (APC) systems have played an increasingly important role in determining the number of passengers in public transport. They are used in the daily monitoring of operations, in long-term demand planning, as well as in revenue sharing within transport associations around the world, which is in the billions annually \cite{vdv:PresseinformationBilanz2018,vdv:PresseinformationBilanz2019, vdv:PresseinformationBilanz2020,wiki:Verkehrsverbund_Berlin-Brandenburg} by both ticket revenues, and public funds
\cite{wiki:Farebox_recovery_ratio} and directly affects services of general interest. Systematic over- or undercounts, so-called \textit{biases} can thus lead to large unfairness and should thus be as small as possible. The VDV 457 is the prevalent standard that defines how APC systems are officially validated \cite{SiebertEllenberger2019,vdv457_v2_1}.

\input{graphics/apc.tex}
Modern APC systems are based on depth-sensing cameras mounted above doors in busses, trains, trams, and other vehicles of public transport (see Figure \ref{fig_apc}). The recorded 3D videos start with a door opening, passengers board and alight, sometimes pushing and carrying objects along like bicycles and baby buggies, and at the end of the video, the door closes again. This sequence of events is called a \textit{video of a door opening phase} or in the following \emph{video} for short.

\citeauthor{9665722} introduce and evaluate the Neural Automated Passenger Counter (NAPC), a deep neural network based on a long short-term memory (LSTM) \cite{hochreiter1997long} architecture \cite{9665722}. It is capable of learning automatic passenger counting \emph{tabula rasa} from videos and total manual counts of e.g.\@{} boarding and alighting passengers \textit{at the end} of each video. To illustrate this, the information \emph{there have been 2 boarding and 3 alighting passengers} already is a label to an entire video, timestamps are not required. This property lowers the effort in acquiring the labels drastically over approaches like object detection \cite{Redmon_2016_CVPR,bochkovskiy2020yolov4} or even segmentation \cite{long2015fully}. Furthermore, the computational effort of the NAPC is comparably low to the aforementioned approaches. Training can thus be executed on consumer GPUs and inference even on legacy hardware systems, see Section \ref{sec_performance}.

We tried to improve the original NAPC architecture by changing popular hyperparameters like the learning rate, batch size, LSTM layer count and depth. What can be considered as a success is the introduction of \emph{learning rate decay} (LRD): before, we have used a \emph{selection set} to determine a suitable epoch. With LRD, we can pick the last epoch, making the process easier to handle. However, besides LRD, our results stayed inconclusive for months, so there was no improvement. Therefore, we switched our exploration to more fundamental parameters like various random seeds and training data selection. It is known from other fields like operation research \cite{SIEBERT20132251} and recent findings for neural networks \cite{Mehrer2020} as well that such seemingly minor choices can have even more impact than algorithms or other design decisions. Another challenge was the lack of a suitable metric, so over time, we created a new one with a universal approach: the \emph{test success chance}. Finally, we did an evaluation on how \emph{stationary} the NAPC is, determined a limit, and found yet another very universal approach to remove it.

We organize the paper as follows: in Section \ref{sec_methods} we introduce the methods we have used. These comprise efficient labeling, LRD, grid search, quantization, ensemble building, and simulative metrics. Then, we present challenges and our results in Section \ref{sec_results}. Finally draw a conclusion in Section \ref{sec_conclusion}.
\subsection{Previous Work}
Over the recent years, together with our research partners, we have developed multiple passenger count algorithms, e.g.\@{} \cite{mdeArticle}, which quality-wise was comparable to non-neuronal methods of other manufacturers. \citeauthor{9665722} introduced the NAPC \cite{9665722} to predict the boarding and alighting passengers on the \textit{Berlin-APC}~\cite{11303_13421} dataset. It consists of roughly 13.000 videos of door opening phases and their total boarding and alighting passenger counts at the video end. A sequence-to-sequence LSTM model predicts the current absolute count for every frame (ranging from 56 to 3275 frames per sequence at 10 frames per second), which means that the prediction on the last frame is trained to match the label and thus is solely used to calculate the accuracy metric. Since the labels are counts, they are all integral, and thus the NAPC predictions are rounded as they typically have decimals. A special bounding box loss function covers all intermediate predictions. The maximum amount of passengers in a \emph{class} (here: boarding or alighting adults) is 67. However, the dataset contains mostly labels with low counts, and videos with high counts are scarce (for details, see Table~\uproman{1} within "Number of Events" in \cite{9665722}).

The NAPC architecture consists of an input and output fully connected (FC) layer and an LSTM core in between (for details, see Figure 2 in \cite{9665722}). Therefore, it does not explicitly use spatial information and interprets each video frame as a flattened 500-dimensional vector. The output is a regression of the cumulative counted passengers until the corresponding frame. \citeauthor{9665722} report a maximum overall accuracy of \textasciitilde 97\% dropping to 55.17\% with their best model in \emph{more difficult} situations like "More than 20 boarding passengers" \cite{9665722}. The global relative bias $\Delta^{(\mathrm{global})}$ and an equivalence test for a 95\% confidence interval are used as an additional metric, fitting the passenger counting task. This is the same methodology as in the recent VDV 457 v2.1 \cite{vdv457_v2_1}, which, however, requires 99\% for high precision systems.
\subsection{Related Work}\label{sec_related_work}

The NAPC as introduced in \cite{9665722} can be considered as a \emph{stationary method} operating on \emph{non-stationary data} since its counts are bounded (see Section \ref{sec_results}) but the labels can basically be arbitrary large. For time series, methods to change non-stationary to stationary data, two popular methods can be applied: The forward difference operator, which is defined by \citeauthor{suwan2018monotonicity} as
\begin{equation*}
\Delta_{h} f(t) = \frac{f(t+h) - f(t)}{h}
\end{equation*}
and the backward difference operator as
\begin{equation*}
\nabla_{h} f(t) = \frac{f(t) - f(t-h)}{h}
\end{equation*}
analogously for the time scale $ h \in \left] 0, 1 \right]$. According to \cite{suwan2018monotonicity} it was first proposed by \cite{miller1989fractional} in 1989. The cumulative sum over a sequence, the approach we use, can be considered the inverse operation of the backward difference operator. The differentiation and integration approach is applied, e.g.\@{} to make stock-market predictions \cite{hillmer1982arima}. In economics, autoregressive integrated moving average (ARIMA) models are used to make time-series forecast predictions on non-stationary data (for details, see Chapter~4 of \cite{box2015time}). The integration step is a cumulative sum over all previous predictions (i.e., an integration over a stationary process). Autoregressive (AR) and moving average (MA) properties help the model to predict future time-steps based on previous predictions (including the errors of previous predictions). Models with those features but no integration are categorized as ARMA. The integration step removes the trend of the time series.

Another topic is the use of legacy computational platforms, which we use for inference: since some have limited floating-point capabilities, we have developed a \emph{universal post-training Monte Carlo quantization method}, which can, in particular, be applied to LSTMs. The term \emph{quantization} refers to the process of reducing the resolution from floating-point to fixed-point numbers while retaining an overall similarity of the values. \citeauthor{dundar1995effects} \cite{dundar1995effects} investigated the effects of multilayer neural network quantization and concluded that 10 bits are necessary for their experiments. However, they used only weight quantization on linear-only connections. More than two decades later \citeauthor{alvarez2016efficient} propose their schematics to quantize recurrent neural networks down to 8-bit resolution \cite{alvarez2016efficient}. They leverage quantization-aware training procedures, allowing for reduced quantization errors post-training. They successfully report the quality of their method with wider LSTM layers (compared to \cite{9665722}). \citeauthor{li2021quantization} proposed a more recent solution to LSTM quantization with and without quantization-aware training in \citeyear{li2021quantization} \cite{li2021quantization}. The authors reduce the resolution for weights and activations to 8-bit fixed-point arithmetics and maintain a 16-bit fixed-point when necessary (e.g.\@{} cell states). Not only is quantization relevant to legacy platforms, but it can also be used to fully utilize specialized neural network ASICs (Application-Specific Integrated Circuit) like the Google Edge TPU \cite{gholami2021survey}.

Another aspect we cover is ensembles: To improve equivalence test success chances by reducing bias, we have combined multiple networks into \emph{NAPC ensembles}. Ensemble methods are used to improve the quality of a prediction of a single \emph{expert model} through the combination of multiple \emph{experts}. Those members of an ensemble then vote, or their predictions are combined such that the resulting selected prediction is a collective solution. Multiple experiments \cite{krizhevsky2012imagenet,qiu2014ensemble} show that ensemble methods can improve the predictions. Keeping the ensemble size small is another objective. Since each member processes the input, the computations grow with each additional member. \citeauthor{zhou2002ensembling} mentions that using all available instances (neural networks in our case) could also result in a worse ensemble performance than choosing a subset of those \cite{zhou2002ensembling}. \\
Multiple learners can be created by e.g.\@{} \emph{Bagging}, \emph{Boosting}, and \emph{Stacking}, which have different properties and limitations. Bagging (bootstrap aggregating) averages through all predictions in regression, or uses plurality voting for classification tasks \cite{breiman1996bagging}. All members are trained independently and can therefore be parallelized. On the other hand, Boosting optimizes recursively and is not parallel. The Adaptive Boosting (AdaBoost) algorithm emphasizes the wrongly predicted samples of one model to train another model with those. The collection of those repeatedly optimized models is weighted and then combined \cite{freund1997decision}. \citeauthor{wolpert1992stacked} introduced \emph{stacking} (or \emph{stacked generalization}) in 1992 as a method to combine the output of one or more \emph{generalizers} \cite{wolpert1992stacked}: \emph{Level 0 algorithms} are optimized on input data and \emph{level 1 algorithms} combine level 0 algorithm outputs to create its own outputs. More levels can be defined analogously. \\
Another aspect of ensemble building is the pruning method to limit the number of learners. Those methods can be categorized into \emph{Ranking}, \emph{Clustering}, \emph{Optimization} and others \cite{tsoumakas2009ensemble}. Ranking-based methods use measurable quality criteria to rank all the available algorithms and take the $k$ best performing to build the ensemble (with size $k$). Clustering-based methods do not need a labeled pruning set as the other methods do, as it clusters the models based on a distance measurement (could be done on artificial data) and prunes the clusters until $k$ members are left. Optimization-based ensemble pruning methods like \emph{Centralized Objection Maximization for Ensemble Pruning} (COMEP) by \citeauthor{bian2019ensemble} starts with a single ensemble member and optimizes an objective iteratively for each new member \cite{bian2019ensemble}. Other methods like genetic algorithms \cite{zhou2003selective} can be used as optimization-based ensemble pruners \cite{tsoumakas2009ensemble} as well.

During our work, another issue emerged: Non-determinism. To conduct research on and potentially exploit the impact of neural network initialization or the training order for the NAPC as in \cite{SIEBERT20132251}, \emph{determinism} is required. This has been an issue with GPU accelerated training, which relies on modern research in the area. As stated by \citeauthor{morin2020non} even with fixed seeds, a robust reproducibility of experiments used to be not achievable \cite{morin2020non}. The non-determinism they measure (variance after training with the same seeds) is only traceable to the GPU floating-point operations, as the initialization of the ResNet~\cite{he2016deep} weights and the order of samples (mini-batches) is fixed. After training 50 models with the same and different seeds, they measure the proportion of variance in the metrics introduced by the GPUs. For the standard deviation of the model's accuracy, the GPU non-determinism contributed around 74\% of the overall variance on the test set, and for the loss, it is more than 87\% of the standard deviation's variance on the test set. However, we could utilize more recent developments that enable deterministic GPU accelerated training \cite{Riach2019}.
\subsection{Contributions}
This work extends the original NAPC with several engineering and research optimizations:
\begin{itemize}
  \item efficient labeling
  \item post-training quantization, including LSTMs
  \item introducing LRD to the training
  \item cumulative summation in training and inference
  \item test success chance as metric obtained by simulation
  \item a grid search to identify important parameters
  \item increased accuracy and less bias by ensemble building
\end{itemize}

%% file: graphics/apc.tex
\begin{figure}
	\resizebox{\columnwidth}{!}{
    \includegraphics[height=\columnwidth]{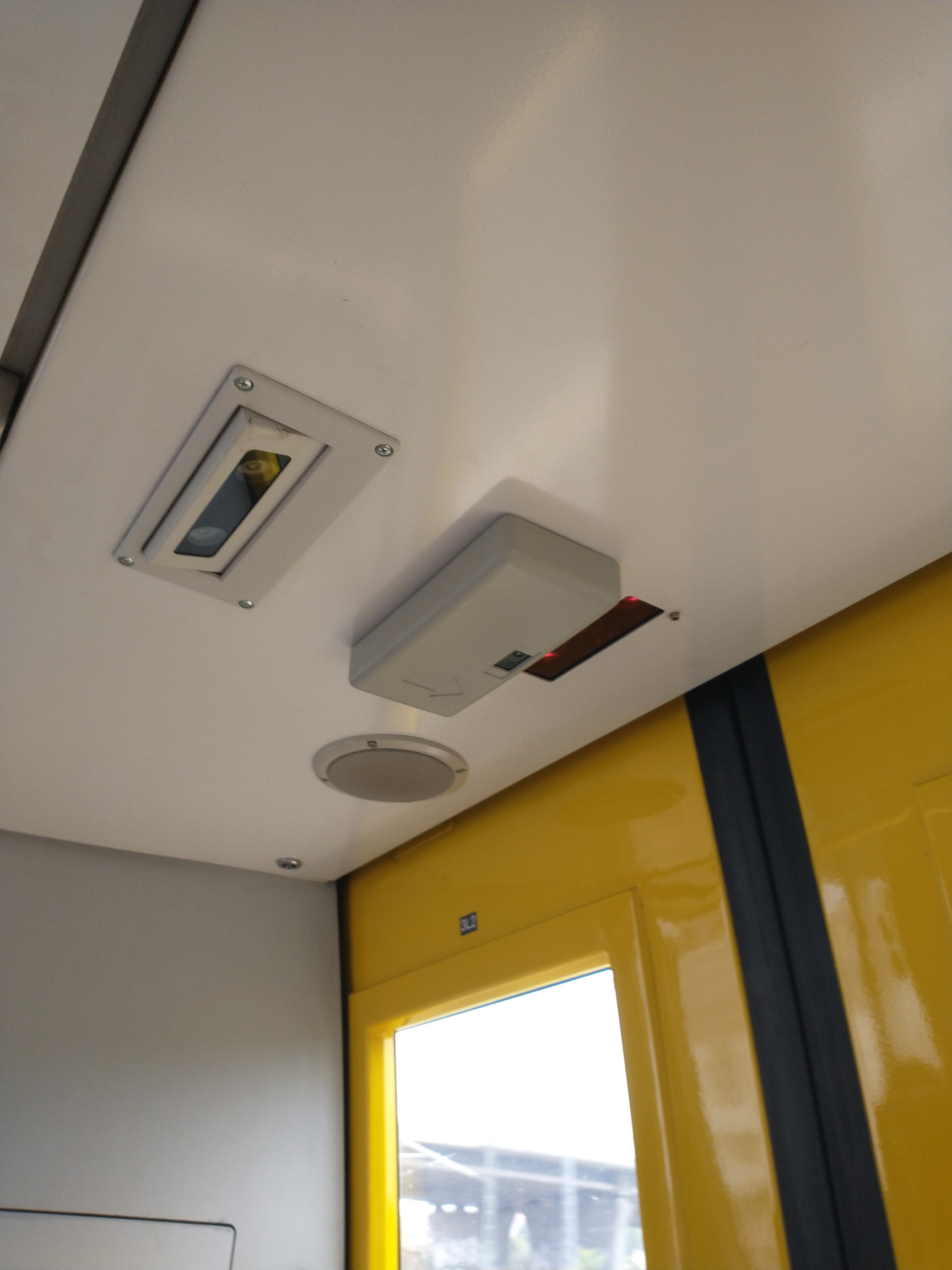}
    \includegraphics[height=\columnwidth]{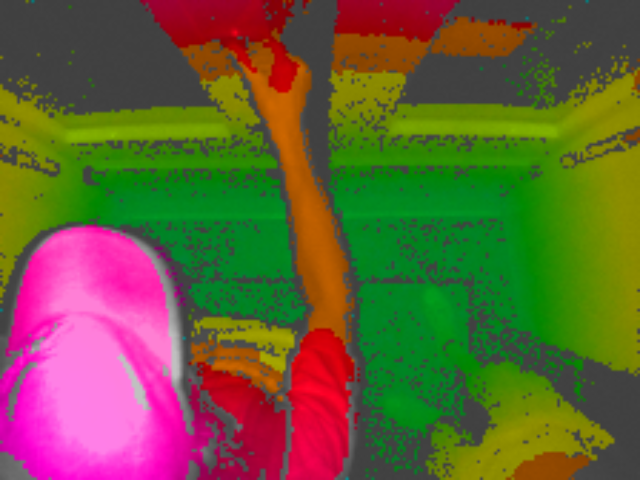}
    \includegraphics[height=\columnwidth]{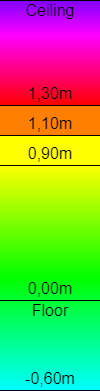}
    }
    \centering
    \caption{Left image: Two automatic passenger counting systems mounted above
    doors, a permanently installed stereographic camera to the left, and a
    temporary installed, battery-powered system with a LiDAR/ToF camera to the right. Center image: a person is triggering a door opening as
    seen from the LiDAR system. Right image: a floor height color scale for the center
    image.}\label{fig_apc}
\end{figure}

%% file: chapters/methods.tex
\section{Methods}\label{sec_methods}
We use the network architecture mentioned in \citeauthor{9665722}
 as baseline predictor with 5 LSTM layers with 50 LSTM units each \cite{9665722}.
 The following methods presented in the subsections are partially combined during the experiments. We did not use the network optimizations from Section~\ref{network_optimizations} within the grid search (Section~\ref{sec_grid_search}). We leverage mixed-precision throughout the training supplied by the Tensorflow framework~\cite{micikevicius2017mixed, abadi2016tensorflow}.
\subsection{Labeling}\label{sec_labeling}
\begin{figure} %
    \includegraphics[width=\columnwidth]{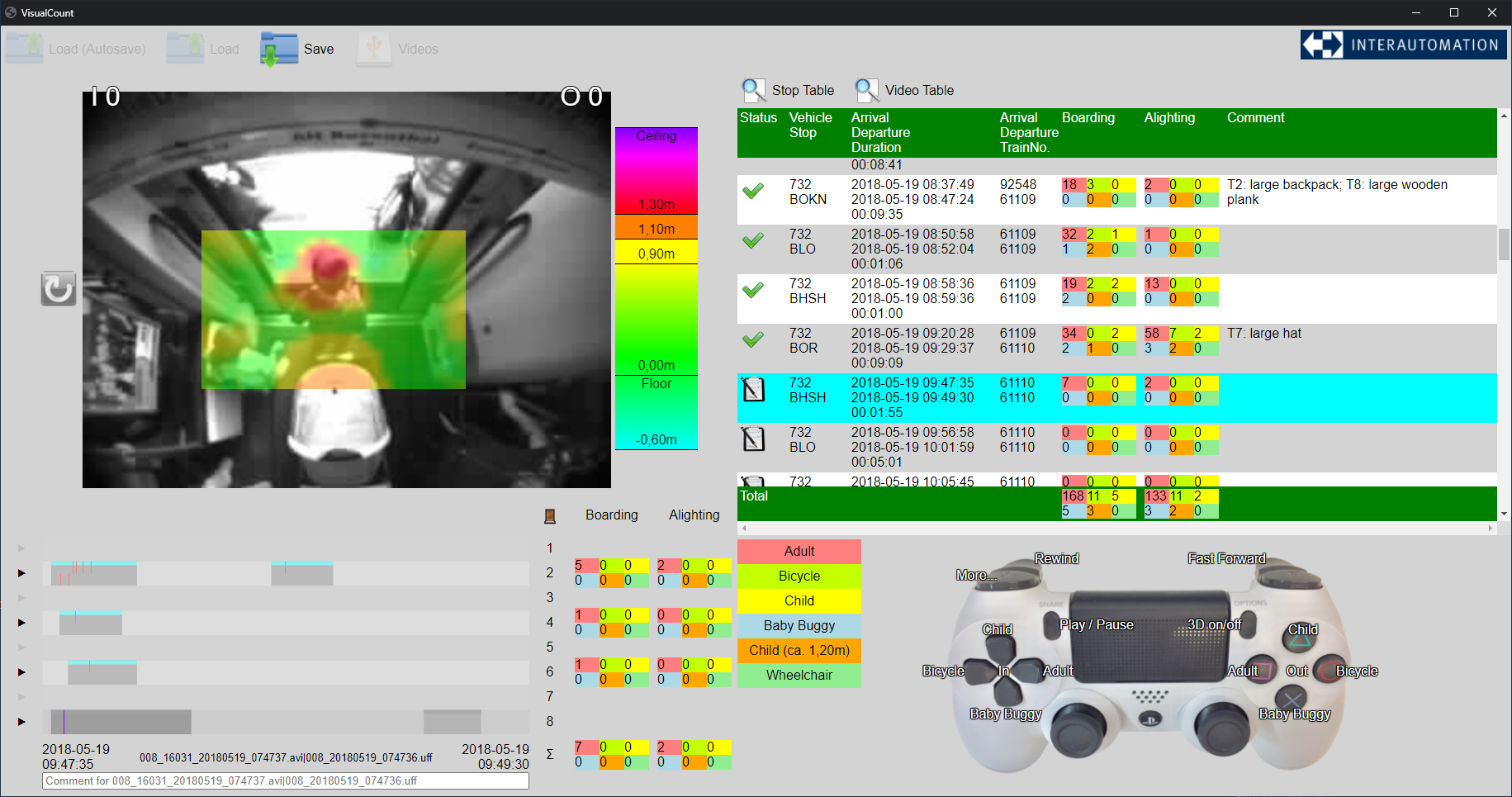}
    \centering
    \caption{VisualCount: A specialized video software to perform manual counts, which we have used to obtain our count data. The height information from 3D videos can be used to distinguish between children and adults, e.g.\@{} by using a critical height of 1.20 meters. To these standards, the tiny person in the center of the image above would be considered an adult, while the one to the left is a child. In the case of our specialized software, we have achieved a 70\% speedup over the use of a standard video player with a spreadsheet program. It runs entirely in the browser; thus, it can operate without a server by using videos from the user's filesystem, which increases data privacy. Users utilize game controllers to navigate through the video with playback or rewind speeds corresponding to the pressure applied to the analog buttons to obtain both slow-motion for crowds and high speeds (10x and more) to skip through long idle timespans. Thanks to a passthrough on filesystem call level, video loading times are typically seconds, even for hundreds of gigabytes of video data.}
    \label{fig_visual_count_screenshot}
\end{figure}

To efficiently gather the large amounts of labels required to train NAPCs, we have developed \emph{VisualCount}, a specialized tool that utilizes game controllers. The main advantage is the analog buttons that allow navigating through a video in both directions at the most appropriate speed: slow in crowds, fast in idle situations, and backward to quickly review complex scenes. It also allows to enable and disable a 3D height map overlay. This allows to distinguish between children and adults by using a certain height threshold (e.g.\@{} 1.20m), see Figure \ref{fig_visual_count_screenshot}. At time of writing, around half a million manual video counts have been made with VisualCount.
\subsection{Network Optimizations}\label{network_optimizations}
The NAPC performs weaker in more challenging scenarios (see Figure 6 in \cite{9665722}) and in particular for high boarding and alighting counts. Predicting non-stationary data (without trend estimation or elimination) is often avoided in regression (compare ARMA to ARIMA in Section \ref{sec_related_work}) by converting it to stationary data. However, this is not possible in the case of NAPC since labels always refer to a range of video images in which the event (e.g.\@{} boarding passenger) happens. Therefore, instead of modifying the data, we modify the neural network and add a cumulative sum at the end, the so-called \emph{cumsum layer}: Before, for each video image, the LSTM needed to report the total count until that image. With the cumsum layer, since frame-wise outputs are summed up, the LSTM only needs to report a frame-to-frame difference. Therefore, the cumsum also preserves the temporal correlation of single events over multiple frames, and the neural part of the network does not need to perform and understand the concept of \emph{summation} anymore. Instead of changing the data, we thus have improved the NAPC to predict on non-stationary data, compare Figure \ref{fig_cumsum}.
\subsection{Training Optimizations}
\subsubsection{Data Storage}
We have stored all our video data in a raw Numpy \cite{Harris2020} array with an additional file that contains the names and offsets of the individual videos and accessed it via a simple Python class. The advantage of this bare-metal approach is that no notable data parsing delays occur once the Numpy file is created, which can be significant for other file formats. Also, we made use of memory mapping, which reduces the data load time basically to zero so code changes can be evaluated \emph{interactively}. Since the Numpy files can become hundreds of gigabytes large, we created them by reading the video files individually and adding the pixel data by appending to the file with \emph{npy-append-array}\footnote{\url{https://github.com/xor2k/npy-append-array}}. Together with memory mapping, this technically very basic approach allows using data sources larger than the main memory of the training machine.
\subsubsection{LRD}
Due to the problem of diverging and collapsing networks during training, we introduce LRD. It decreases the gradient step size over time, stabilizes the learning procedure, and allows us to pick the last epoch of the training run. It eliminates the need to pick an epoch by using a selection set. Therefore, more videos are available to the training and test set as compared to the previous solution.
It was already stated by \cite[][p.20]{bengio2012practical} that the learning rate is one of the essential hyperparameters.
\subsubsection{Grid Search}\label{sec_grid_search}
We wanted to know to what amount similar effects as in \cite{SIEBERT20132251} can occur for deep learning neural networks. In \cite{Mehrer2020}, the authors focus on neural network initialization while we evaluate multiple randomness aspects at the same time and their impact on the quality of the NAPC. We split the data into mutually exclusive groups to measure the impact of the training data on the results. Our grid search covers the product of the following ranges:
\begin{enumerate}
    \item \textit{group selection}: 4 options
    \begin{itemize}
        \item Uniformly shuffling 8000 videos
        \item Dividing into 4 mutually exclusive groups (with index $g$) with 2000 videos each
    \end{itemize}
    \item \textit{data amount}: 2 options
    \begin{itemize}
        \item 2000 training videos, selected by group index $g$ (smaller training set)
        \item 6000 training videos, everything but group index $g$ (larger training set)
    \end{itemize}
    \item \textit{weights random seed}: 10 options
    \begin{itemize}
        \item Affects only the initialization of the weights
    \end{itemize}
    \item \textit{training random seed:} 4 options
    \begin{itemize}
        \item Affects order of training videos in the mini-batches (and the concatenation of the videos)
        \item Affects dropout regularization~\cite{srivastava2014dropout} (i.e., the random masking of values)
    \end{itemize}
\end{enumerate}

\subsection{Inference Optimizations}
\subsubsection{LSTM Quantization}
\citeauthor{zhu2020towards} pointed out that the weights of artificial neural networks are trained as floating-point numbers since they allow for higher precision and more stable learning \cite{zhu2020towards}. Previous work on post-training quantized convolutional neural networks revealed the small impact of lower precision weights on the prediction quality. Major frameworks such as Tensorflow \cite{abadi2016tensorflow}, and PyTorch \cite{paszke2019pytorch} have build-in tools to reduce the machine precision of the trained weights and to quantize them to fixed-point numbers. However, LSTMs are not well supported or in a beta state. As they are the core element of the NAPC, quantization is needed to perform faster inference. Our custom framework approach is a greedy Monte Carlo random search of input and weight scales that processes each layer in sequential order. This approach reduces the computational effort and time required instead of optimizing all layers at once. The quantizer uses several hand-picked calibration videos (less than a dozen) as a reference, and any quantized model is required to predict within an error margin $\varepsilon$ for every timestep. No labels are required for this process. Also, we have kept the reference video set intentionally small. A benefit is not only a fast quantization process but also an \emph{increase in variability}: we only require the quantized models to be within an error margin for a few videos and thus only relatively \emph{loosely related} to the non-quantized model. We treat a quantized model as if it had been an \emph{entirely unrelated} to its non-quantized counterpart, so it undergoes full validation just like any other NAPC. This approach allows us to increase our training pool size without training more models. \\
From a structural point of view, apart from the activation functions, the network consists of two FC layers and an LSTM core. In the following, variables with $\widetilde{\cdot}$ are quantized and therefore scaled versions, and $\times$ denotes the element-wise product. Scalars are denoted with lower case letters in the following equations, while vectors and matrices are denoted with upper case letters. \\
FC layers are stateless, and thus the layer indexing does not consider timesteps but only placement within the network. Such layer with index $n$ receives the scaled output $
\widetilde{O}_{n-1}$ of the previous layer $n-1$ as input, as well as the normalization factor $o_{n-1}$. Such that
\begin{align*}
	O_{n-1} =  \frac{\widetilde{O}_{n-1}}{o_{n-1}} - \varepsilon
\end{align*}
as $\widetilde{O}_{n-1}$ is the scaled and quantized version of $O_{n-1}$ which could have rounding errors $\varepsilon$. When searching for appropriate input and weight scales ($i_{n}$ and $w_{n}$ respectively), we have to scale the input $\widetilde{O}_{n-1}$, weights $W_{n}$ and biases $B_{n}$ for layer $n$ accordingly. Therefore, we calculate the output $\widetilde{O}_{n}$ and its scale $o_{n}$ as follows:
\begin{align*}
\widetilde{I}_n ={}& \frac{\widetilde{O}_{n-1} i_n}{o_{n-1}} \\
\widetilde{W}_n ={}& W_n w_n \\
\widetilde{B}_n ={}& B_n w_n i_n \\
\widetilde{O}_n ={}& \widetilde{I}_n \widetilde{W}_n + \widetilde{B}_n \\
o_n ={}& w_n i_n\ \ .
\end{align*}
We extend the previous notation to the stateful case of LSTMs with the timestep $t$ superscript. Again we have the output $\widetilde{O}^{t}_{n-1}$ of the same timestep, but the previous layer. Additionally, for LSTMs there are the hidden $\widetilde{H}^{t-1}_{n}$ and cell $C^{t-1}_{n}$ state of the same layer of the previous time step, where the first one is scaled with $i_{n}$ and the latter one is always unscaled (i.e., floating-point). We define $M[\{i,f,o,c\}]$ as the non-activated floating-point LSTM gate block matrix, which is calculated with the additional recurrent kernels $U_{n}$, as well as the weights $\widetilde{W}_n$, and biases $\widetilde{B}_n$; which are block matrices themselves with the same indices. Furthermore, $\sigmoid(\cdot)$ is the logistic sigmoid activation function. Those definitions lead to the following calculations:
\begin{align*}
\widetilde{I}^t_n ={}& \frac{\widetilde{O}^{t}_{n-1} i_n}{o_{n-1}} \\
\widetilde{W}_n ={}& W_n w_n \\
\widetilde{U}_n ={}& U_n w_n \\
\widetilde{B}_n ={}& B_n w_n i_n \\
M[k] ={}& \frac{\widetilde{I}^t_n \widetilde{W}_n[k] + \widetilde{H}^{t-1}_n \widetilde{U}_n[k] + \widetilde{B}_n[k]}{w_n i_n}, \\
& \text{for all } k \in \{i,f,o,c\} \\
C^t_n ={}& \sigmoid \left( M[i] \right) \times \tanh \left( M[c] \right) \\
	& + \sigmoid \left( M[f] \right) \times C^{t-1}_n \\
\widetilde{H}^t_n ={}& i_{n}(\tanh \left( C^t_n \right) \times \sigmoid \left( M[o] \right)) \\
\widetilde{O}^{t}_{n} ={}& \widetilde{H}^t_n\\
o_n ={}& i_n\ \ .
\end{align*}
As $\widetilde{H}^t_n$ is also the input for the next layer and timestep, it is multiplied again with $i_{n}$.
\subsubsection{Ensemble Building}
Ensemble methods are commonly used in machine learning to overcome the weaknesses of a single predictor. We opt for ranking-based pruning \cite{tsoumakas2009ensemble}, where we sort the independently trained networks according to the probability of passing the VDV 457 v2.1 and select the top $N$ networks. The final prediction of the ensemble is determined via the $\tau$-th percentile or \emph{quantile} for each class, which is particularly well suited to compensate for a bias and can be tuned via a calibration set. This process is a stacking ensemble method as we optimize a meta-model (quantile regression) based on the specific ensemble. In the following, unless we combine ensembles specifically for different training (sub-)datasets, the only differences between most ensemble members are solely their random seeds. Also, note that, from a mathematical point of view, the seemingly familiar and straightforward median-like functions like quantile are highly non-linear, non-differentiable, and thus not trainable by gradient descent. Therefore, they structurally provide added value to typical neural network training.
\subsection{Simulative Metrics}
Deep Learning uses gradient descent and variations of it, which requires a \emph{differentiable} loss function. Accuracy, while not being differentiable, is constantly measured and reported during the training process and is a key metric. However, for automatic passenger counting, accuracy is not well suited since it is very much decoupled from the bias: an NAPC instance can have an extremely high accuracy (e.g.\@{} 98\%), but the bias might vary from 0\% to multiple percents because the few videos where the model's prediction does not match the label (e.g.\@{} the remaining 2\%) typically contain difficult situations with crowds and thus the neural network makes a large error. On the other hand, to pass the VDV 457 v2.1 or the equivalence test, the standard deviation is important as well. Our initial idea was to use some linear combination of the absolute bias and standard deviation. However, this approach is quite arbitrary, and not less complex as doing an entire equivalence test in the first place. The drawback is that the sample size cannot be controlled, and the possible outcomes are only whether the test was passed or failed with nothing in between. Labeling tens of thousands of videos to run multiple equivalence tests and average over tests would be affordable thanks to our efficient labeling software mentioned in Section~\ref{sec_labeling}, but it still scales badly. Therefore, as called \emph{test success simulation} in \cite{EllenbergerSiebert2021}, we use another approach: we assume that the error made on our validation set generalizes well enough so that it can be used as an \emph{empirical} error distribution to sample from \cite{vaart_1998}. We then draw such a sample of the desired sample size (e.g.\@{} 3600 videos) and repeat the process (e.g.\@{} 10000 times), and determine the \emph{test success ratio} or \emph{test success chance}; which is a so-called \emph{bootstrapping} method \cite{osti_5421967}. Another advantage of this approach is that no mathematical model for the error distribution with many incomprehensible parameters to tweak is required, and multiple categories like boarding and alighting passengers can be meaningfully combined into one continuous metric.

%% file: chapters/results.tex
\section{Results}\label{sec_results}
\subsection{Network Upper Bounds}
Investigations revealed a limitation of the architecture from \cite{9665722} (see Figure \ref{fig_cumsum}): the original NAPC (without cumsum) cannot count above a threshold of around $100$ passengers, which can easily be observed by looping sequences multiple times, e.g.\@{} a sequence with one boarding passenger 200 times. The training of models with cumsum leads to unaffected predictions during inference \emph{regardless} of the number of passengers. However, this is not the only issue: networks without cumsum also suffer from random predictions of the opposite class for sequences with more than 70 passengers.

Even though LSTMs are capable of learning non-linearities in general and should therefore be able to predict non-stationary data, our experiments have shown that there are limits. When applying cumsum to our model, the LSTM is not required to learn and perform the arithmetic logic of counting (summation) and thus does not need to compensate for the stochastic drift induced by rising passenger counts anymore. Cumsum layers however, as can be seen in Figure \ref{fig_napc_cumsum_models}, come at a cost: the average model quality gets worse. Therefore, training more neural networks (e.g.\@{} with different random seeds) is required to obtain a NAPC pool with comparable success chance. 
\input{graphics/cumsum.tex}
\subsection{Fixed Training}
Initially, we could not train deterministically, even after setting all seeds. Determinism within the framework, drivers, and hardware is required to e.g.\@{} research the influence of random seeds and fortunately became available during our research \cite{Riach2019, nvidia2021determinism}. As it turns out, the framework we used (TensorFlow) did not support deterministic training out-of-the-box (without patching) before v2.3 \cite{nvidia2021determinism}. From that version on, it could be activated programmatically. Further, an issue in one of the GPU libraries (cuDNN) affected determinism in LSTMs, which was fixed (for details, see \cite{cudnn_v7_6_1_documentation}) around the time we were about to start the experiment.
There are other factors limiting the reproducibility of deep learning in general \cite{alahmari2020challenges}:
\begin{enumerate}
    \item \textit{CPU architecture}
    \begin{itemize}
        \item E.g.\@{} whether it is x86 or ARM
    \end{itemize}
    \item \textit{GPU manufacturer and GPU hardware architecture}
    \begin{itemize}
        \item E.g.\@{} whether a Nvidia Turing (RTX 20XX) or Ampere (RTX 30XX)
        was used
    \end{itemize}
    \item \textit{Any software and driver version \& build}
    \item \textit{The operating system version \& build}
\end{enumerate}
It turned out that, on top of fixed hardware, having fixed software versions may suffice for determinism, but not for reproducibility: using the same framework version, but from different sources (PIP, Conda, compiled from source) resulted in different randomness and thus different, incompatible training results. Therefore, we eventually used a TensorFlow docker image which can be fixed by its hash, and used only one machine equipped with four identical GPUs.
\subsection{Grid Search}\label{sec_res_grid_search}
\input{graphics/grid_search_epoch.tex}
\input{graphics/grid_search_sample_size.tex}
\input{graphics/anomaly.tex}
Since variation among trained NAPC models was generally very high and we were thus unable to reliable judge whether a modification to the architecture or hyperparameters improved quality or deteriorated it, we investigated a more fundamental aspect of training: randomness. Unlike \cite{frankle2018lottery, Dick2014HowMR} we decided not to fix random seeds but iterate over multiple. Such experiments can pinpoint a random variable like the initialization of the neural network or the data training order and its effect on the overall training. From other domains like operation research, it is known that randomness can have a far more significant influence than other popular design aspects \cite{SIEBERT20132251} and has recently been shown for deep learning as well \cite{picard2021torchmanualseed3407}. Therefore, to achieve a minimal level of conclusiveness, iterating over multiple seeds is necessary and can be considered essential for research in general \cite{hutson2018artificial, boulesteix2020replication}.

There are several kinds of randomness during the training process: Firstly, the selection of the subset of data used for training\footnote{\citeauthor{9665722} have done this reproducibly, and independent of the framework in \cite{9665722} already.}, secondly the random initialization of the weights, and thirdly, the randomness during the training, which spans from the data preprocessing and augmentation, to also regularization methods.
For a visual overview, compare Figure \ref{fig_grid_search_epoch} and Figure \ref{fig_grid_search_sample_size}. The most obvious observation is that more training data (6000 vs.\@{} 2000 videos) increases test success chance. The qualitative difference among the models can be very high, which needs to be accounted for when making decisions about deployment or further development, even without non-determinism introduced by GPUs. Building (unranked) ensembles increase the success chances if a quantile > 50\% is chosen as it compensates undercounting, which the NAPC method seems to be more prone to than overcounting. Even when all 8000 possible training videos are available to the ensemble via its members, the count quality is still higher if every ensemble member receives more training videos, so generalization is favorable over specialization. There seems not to be a big difference in the prediction quality introduced by the randomness in initializing weights and training (dropout layer and composition of batches). However, some configurations lead to a terrible count quality even with all training videos being available to a bias-corrected ensemble (see reddish highlighted subplot in Figure \ref{fig_grid_search_sample_size}). This is an interesting and unexpected case: when combining a certain neural network initialization seed with a certain training order seed, \emph{no matter the data}, the training is almost certain to fail. The mentioned seed combination would be a starting point for further investigation and could lead to a better architectural understanding of the neural network.
\subsection{Inference}
\subsubsection{Quantization}
We could almost entirely quantize our network, except for the non-linear LSTM activation functions. Related work \cite{choi2018pact, alizadeh2020gradient, lin2016fixed} leveraged restricted activations, regularization during the training, or quantized training (fine-tuning) to improve quantizability, while NAPC uses neither of these. Compared to feed-forward neural networks, RNNs maintain an internal state. In our case, videos can range from a few dozens to multiple thousands of frames with a large variety of counts and situations. The varying length temporal dimension makes it more difficult to analytically quantize networks as it is done for other architectures \cite{banner2018post}. We quantize all weights, biases, inputs, and outputs to be an integer (i.e.\@{} fixed-point). We defined $\varepsilon$ as the absolute error margin for each frame and class during quantization. It is set to at most $0.45$ absolute deviation from the floating-point NAPC's output as the final predictions are on-average within less than $0.02$ to the closest integer. The idea is to increase quantization potential as much as possible since the quantized model needs to pass the same validation as its non-quantized template model. With this approach, we discourage divergent counts in most cases, resulting in less degraded prediction quality. If the quantized version predicts more than $\varepsilon$ absolute deviation during the process, the scale is discarded, and new scales for this layer are drawn randomly. One could choose $\varepsilon$ to be even smaller as this would result in tighter inference quality, but it also increases the risk of violations during the quantization process and could therefore take more time or could not even finish for some layers. Input and weight scales are optimized concurrently as the algorithm works through each layer.
We confirm that the empirical results of this approach are sufficient to leverage quantization for real-world usage of NAPC in existing edge devices. Both 16 and 32-bit fixed-point quantization are feasible for the greedy approach. Lower precisions were only possible for particular layers. However, those increased the quantization noise for subsequent layers, thus providing no advantage. With a 32-bit fixed-quantization, we can achieve comparable results to unquantized models, with a maximum absolute deviation of 4 to the ground truth (floating-point prediction) sampled from 13 randomly chosen quantized models and checked against all 4966 hold-out sequences (unseen during training and calibration).

In other experiments and to our great surprise, a model quantized with this method is not necessary of inferior or of equal quality as its non-quantized template but can be \emph{superior} as well even to an extent where some mediocre model turns into one of the \emph{best models} after quantization. Therefore, quantizing all models can be a convenient way to enlarge the search space and increase quality. Nevertheless, the average quality of our quantized models is below those without quantization, compare to Figure \ref{fig_napc_cumsum_models}.
\subsubsection{Ensemble building}
\begin{figure}[ht!]
    \includegraphics[width=\columnwidth]{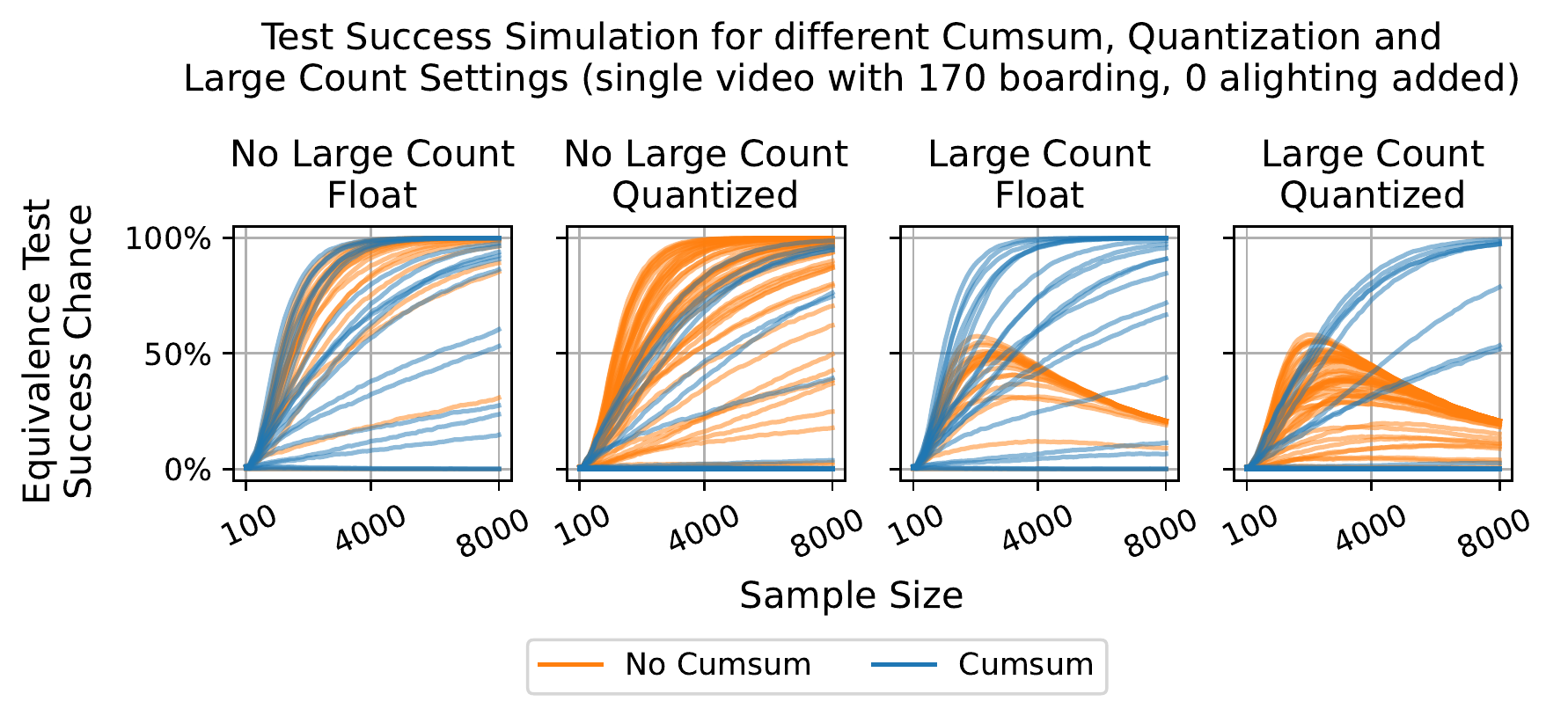}
    \centering
    \caption{Test success simulation, comparison with and without cumsum, quantization, and a single large count video. All NAPCs shown here have been trained using the whole 8000 video training set: even though certain models might improve, quantization reduces the average model quality. Also, in scenarios without large counts, models seem to perform better without cumsum. However, even a single video with many boarding passengers (here 170) can drastically reduce test success chances without cumsum, while NAPC instances with cumsum are mostly unaffected. An explanation for the non-convexity in the two right figures can be found in Figure \ref{fig_anomaly}. Overall, creating many instances improves the chances of obtaining high-quality models.}\label{fig_napc_cumsum_models}
\end{figure}

For the NAPC, we can confirm studies about random seed ensembles that suggest they can improve quality \cite{hansen1990neural, fort2020deep}. 
As mentioned previously, we take the $\tau$-th (with $\tau \ge 0.5$ ) quantile\footnote{The choice $\tau = 0.5$ is also known as \emph{median}.} of the ensemble for each class for the final prediction. This is due to the property of NAPC instances undercounting on average, which is even worse when using the cumsum, compare the lower success in Figure \ref{fig_napc_cumsum_models}. Also, this approach filters outliers and selectively centers the error distribution of the ensemble more towards 0, which increases the test success chance, compared Figure \ref{fig_grid_search_epoch} and Figure \ref{fig_grid_search_sample_size}.
\subsection{Performance}\label{sec_performance}
\begin{figure}[ht!]
    \includegraphics[width=\columnwidth]{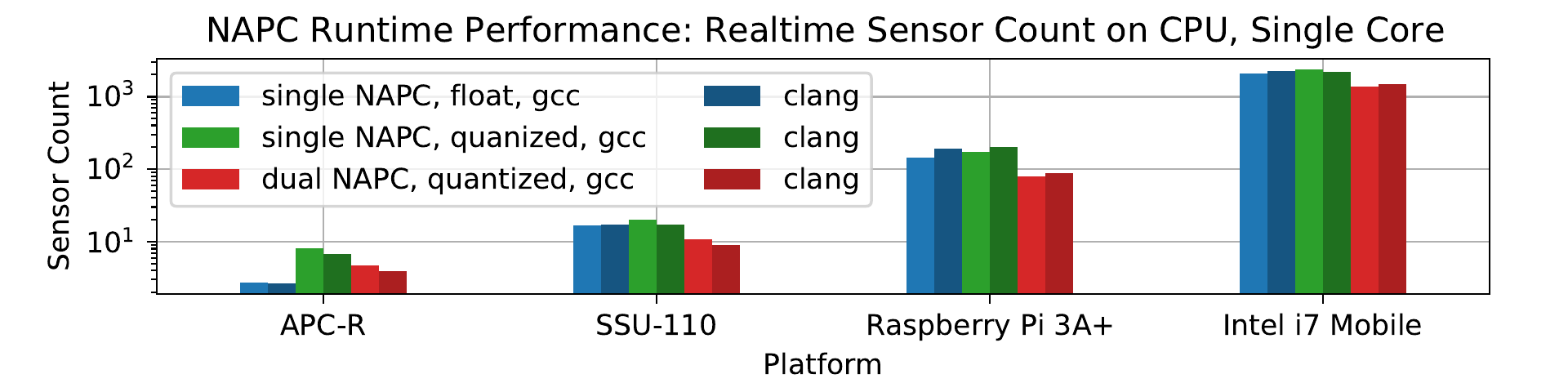}
    \centering
    \caption{%
NAPC performance was measured in real-time sensor counts on different platforms using different compilers. GCC (v11) seems to perform particularly well on legacy platforms, while Clang (v13) performs better everywhere else. The APC-R profits from quantization the most as it does not have a dedicated floating-point unit. A sensor is assumed to operate at 10 FPS.}\label{fig_napc_performance}
\end{figure}
To compare the performance of a quantized and a floating-point NAPC instance, we developed a custom inference framework written in C/C++ \cite{kernighan2006c, stroustrup2000c++}. To increase the performance, we include the weights in the binary and do not use dynamic memory via \texttt{malloc} or \texttt{new}. This allows us to infer multiple sensors concurrently on various platforms (x86, ARMv7, and ARMv5). For the benchmark, we use a video with a length of approx. 204 seconds, which can be stored locally even on a legacy machine with only 32 MB of free storage. We measure the user time to infer all of these sequences to calculate how many NAPC instances the system could run in real-time. In Figure \ref{fig_napc_performance} we show the performance of different CPUs. Noticeably the already deployed SSU-110 and APC-R can reach around 20 (SSU-110) or approx. \@{} 6 (APC-R) NAPC instances in real-time while x86 machines like the Intel i7 can do thousands and even a Raspberry Pi up to around 200 with multicore systems only using a single core. The APC-R profits most from quantization, as it would only compute approx. 2 times real-time with floating-point models.

%% file: graphics/cumsum.tex
\begin{figure}
  \centering
  \includegraphics[width=\columnwidth]{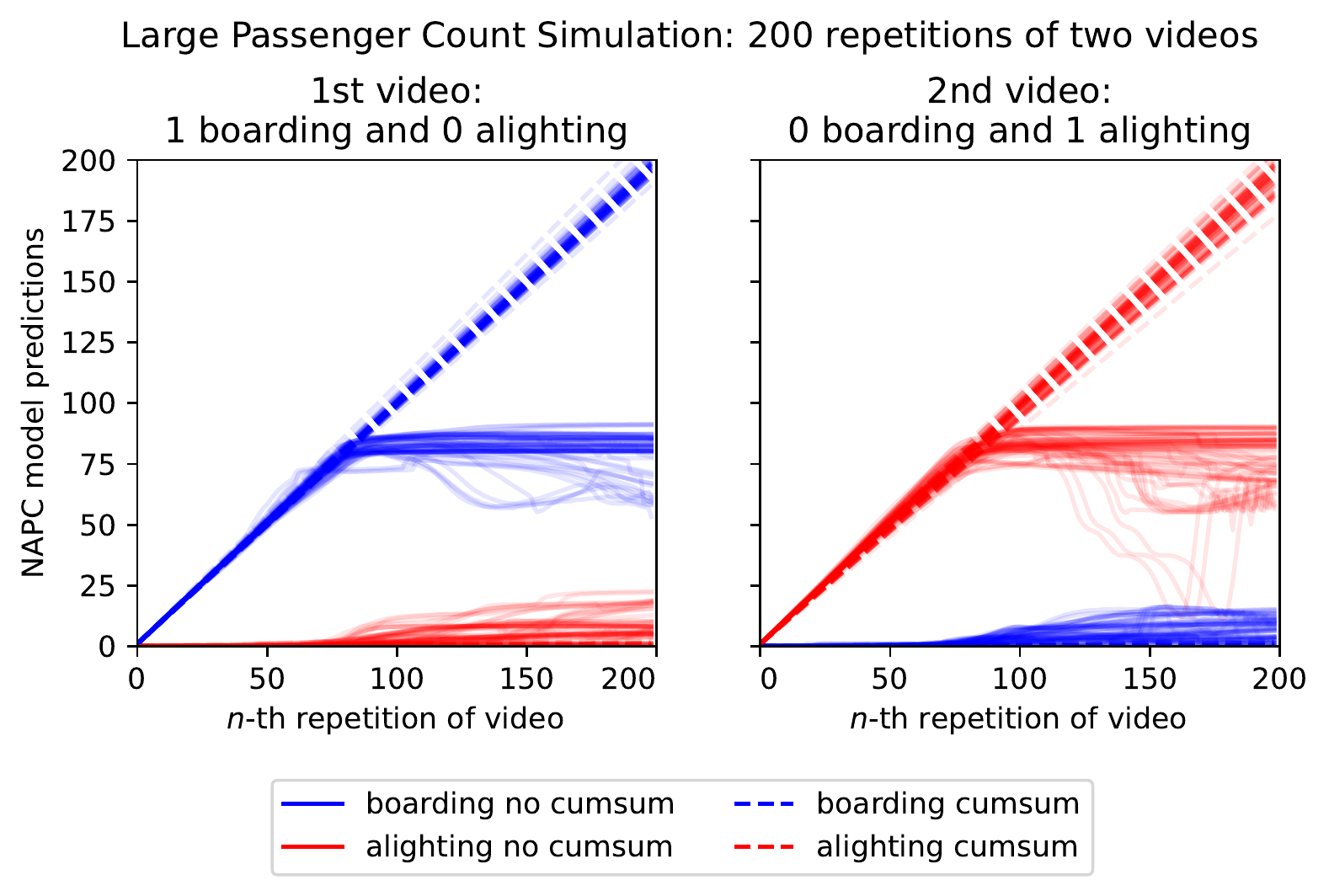}
  \caption{The effect of the cumsum layer: all NAPC instances (floating-point and quantized) from the qualitative performance experiment displayed in Figure \ref{fig_napc_cumsum_models}. Left plot: the expected number of boarding passengers after the last video iteration is 200 and 0 alighting. However, without cumsum (continuous lines), the highest counting NAPC instance has only reached a maximum of 91 boardings (91.43 without rounding) passengers after the last repetition. This can be considered a best-case upper limit, as the NAPC counts vary heavily between individual instances. Although there were supposed to be none for alighting passengers, the NAPC instances have counted up to 23 (22.64 without rounding) in the left plot after the last repetition (worst-case scenario). The NAPC instances with cumsum (dashed lines) counted on average 197 ($197.38 \pm 3.3$) boarding and 0 ($0.32 \pm 0.73$) alighting passengers in the left plot, which almost matches the labels (200 and 0). Additionally, models without cumsum can -- and some do -- count backward after a certain number of video repetitions, while the model with cumsum by design is unable to do so: the absolute activation, which only produces values in $\mathbb{R}^+_0$, happens before the cumulative summation.
  Right plot: the same scenario as with the left plot, based on a video with zero boarding and one alighting passenger, the expected label after 200 iterations would be 0 and 200. As one would expect, the results are very similar: the NAPC instances without cumsum reached up to 15 (15.26) boarding and 90 (90.46) alighting passengers, while the instances with cumsum as used in the left plot counted on average 0 ($0.34 \pm 0.61$) boarding and 193 ($193.29 \pm 5.7$) alighting passengers, which is again a slight undercount.
}\label{fig_cumsum}
\end{figure}

%% file: graphics/grid_search_epoch.tex
\begin{figure*}[!ht]
    \includegraphics[width=\textwidth]{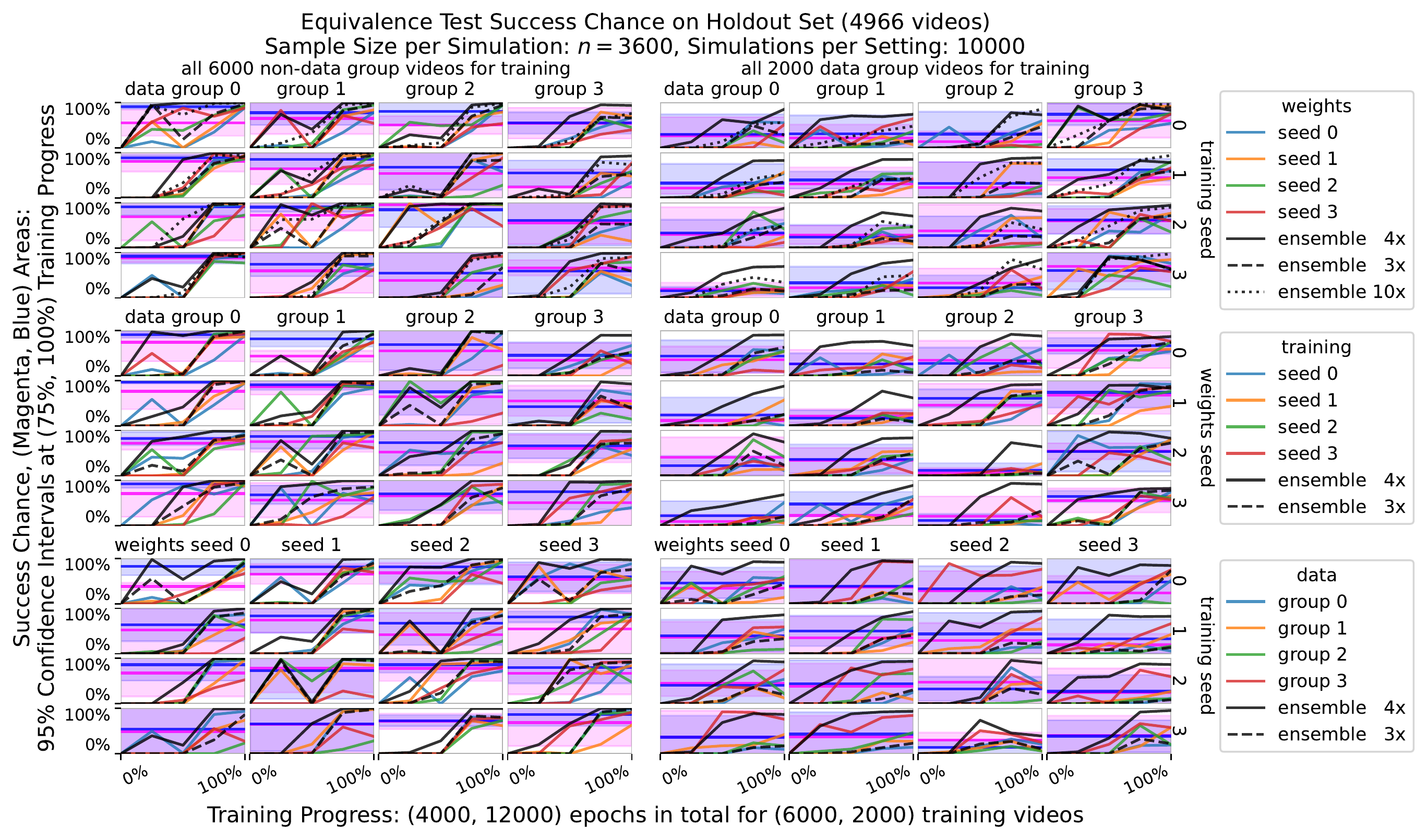}
    \centering
    \caption{Grid search results, equivalence test success chance as a function of training progress (sample size $n=3600$). The training dataset size has the greatest impact: more training data improves the success chance drastically. This means that the capabilities of our small NAPC neural network variant have not yet been exhausted. The spread among the different models can be very high. After the training video count, the data group has the second-largest impact. Although these groups have been chosen as a random division of the full training dataset, data group 3 worsens success chance when absent and improves it when present. The 4x ensemble using a $2/3\approx 66.6\%$ quantile performs better than the best model most of the time. This is most likely due to its ability to reduce undercounting since the other ensembles with 50\% quantile (3x ensemble) or $6/9\approx 55.5\%$ quantile (10x ensemble) yield lower counts on average. In most but not all cases, the last training epoch has the best quality, in particular when training with the larger 6000 video set. For a sample size perspective, see Figure \ref{fig_grid_search_sample_size} for comparison.}\label{fig_grid_search_epoch}
\end{figure*}

%% file: graphics/grid_search_sample_size.tex
\begin{figure*}[!ht]
    \includegraphics[width=\textwidth]{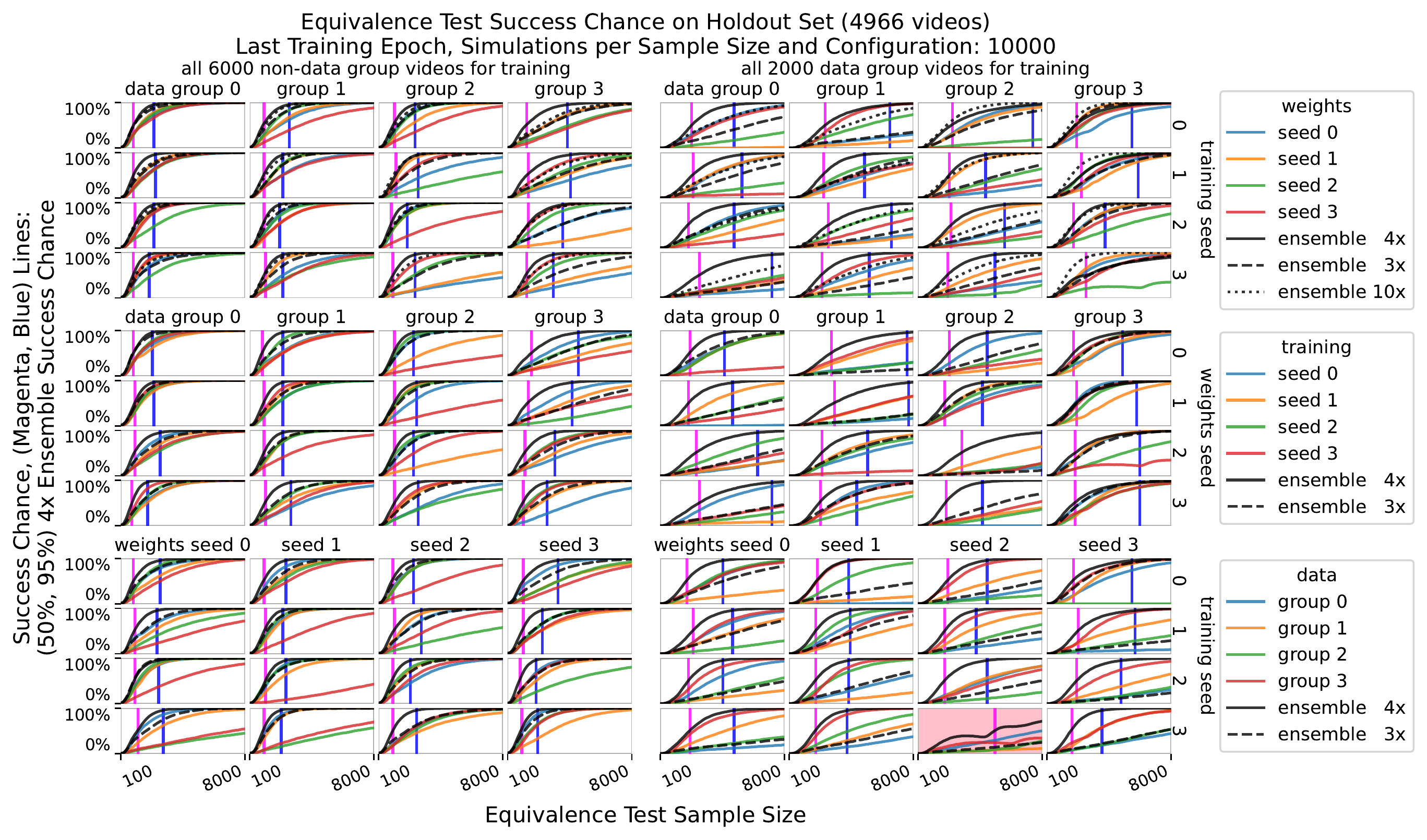}
    \centering
    \caption{Grid search results, equivalence test success chance at 100\% training progress (in contrast to the training process in Figure \ref{fig_grid_search_epoch}). NAPC instances trained with more data tend to require a smaller sample size and thus less effort and cost to succeed in the equivalence test, particularly for reliable success (success chance $\ge$ 95\%). Ensembles perform around as well as the best model. Note that some lines are not convex. This occurs when there is a substantial, single error spike (see Figure \ref{fig_anomaly}).
All in all, NAPC models can be validated safely with samples of around 3600 door opening phases, which is about 60\% less than 6147 stop door events or around 9000 door opening phases as suggested in the VDV 457 v2.1 standard in the case of the vehicle we have used. Latter large sample sizes we have proposed became standard in the pre-NAPC era in 2018 so that the then prevalent legacy APC systems had a fair chance to succeed. The left half of the figure shows that using more training data (generalization) per NAPC training is favorable over the training with less data (specialization) in general and for creating ensembles. We also detected an artifact in the bottom right $4 \times 4$ grid: The reddish highlighted subplot at (training videos, weights seed, training seed) = (2000, 2, 3) indicates that not a single NAPC instance or ensemble would achieve a 95\% success chance. That means, regardless of the data (group) used, with those fixed seeds for the initialized weights and the randomness during the training process, we could not succeed in training an acceptable model, compare Section \ref{sec_res_grid_search}.}
\label{fig_grid_search_sample_size}
\end{figure*}

%% file: graphics/anomaly.tex
\begin{figure}[!ht]
    \includegraphics[width=\columnwidth]{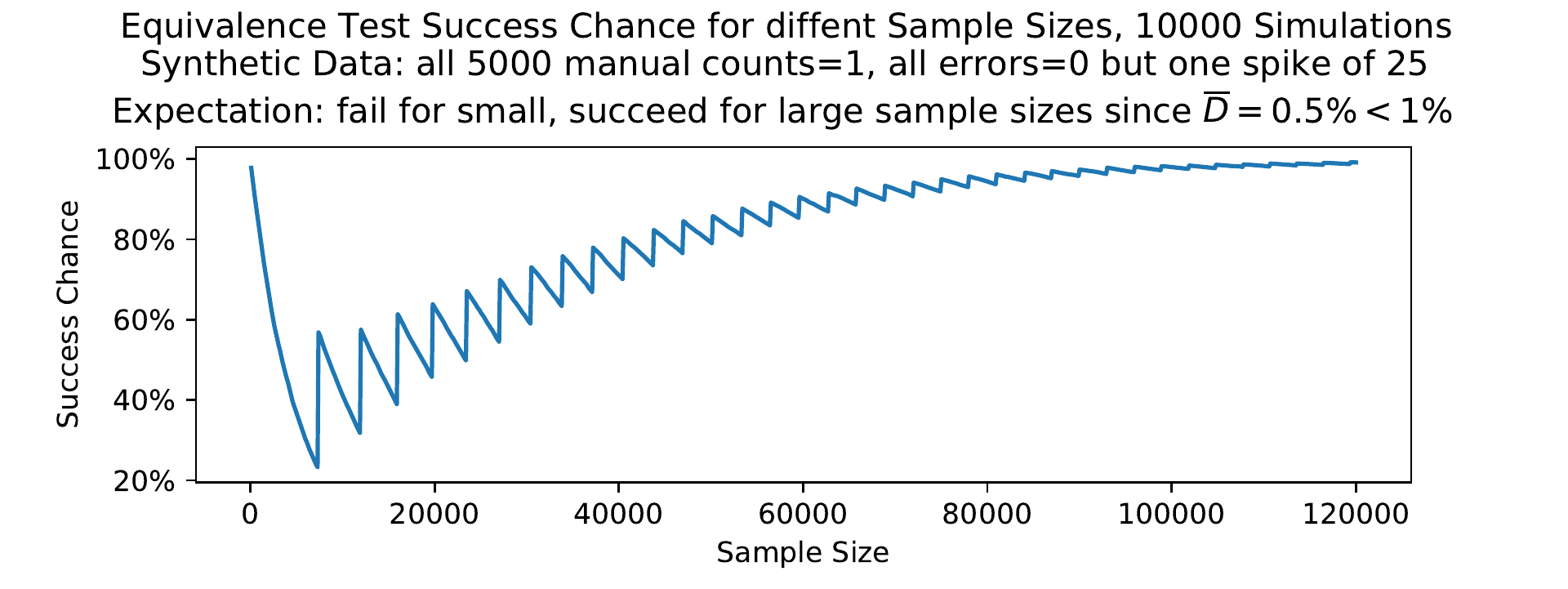}
    \centering
    \caption{In Figure \ref{fig_grid_search_sample_size} some unexpected non-convex curves occur when there is a single substantial error in the population. We extracted this into a synthetic data example. For small sample sizes, chances to hit the error are relatively low, thus the high success chance. As the sample grows, the success chance worsens and increases again as the standard error increases less. A sawtooth-like shape indicates that this property repeats on smaller scales and can be found in real-world examples, although it is less pronounced there.}
\label{fig_anomaly}
\end{figure}

%% file: chapters/conclusion.tex
\section{Conclusion}\label{sec_conclusion}
We proposed several improvements to the training and inference of existing NAPC implementation, which considerably enhance the quality of the predictions and usability in real-world scenarios.

For a long time, it was unclear whether the new VDV 457 v2.1 could ever be passed with 99\% count accuracy\footnote{An accuracy of 99\% refers to the equivalence test for bias, i.e.\@{} the absolute bias being $<1\%$ including confidence intervals.}. Our results prove that it can be done. Thanks to quantization, both in theory and practice, even on legacy hardware already installed in vehicles. Not only can the test be passed, but also the validation costs of this APC system are around 50\% less out of the box due to lower sample size requirements\footnote{3600 videos $\approx$ 3000 stop door events instead of the 6147 mentioned in the VDV 457 v2.1 which we proposed based on low precision legacy systems back in 2018 to give APC manufacturers a fair chance to pass the test.}.

We have created and applied a variety of methods for quality control and identified that the number of videos in the training process matters a lot, and additional measures like \emph{quantile ensembles} are required to compensate for the undercounting that seems to be intrinsic to the NAPC. Further research has to be done to determine the limits of the architecture, e.g.\@{} a maximal count of training videos at which quality does not further improve. With the test success chance, we have introduced a \emph{simulative metric} based on an empirical distribution. Our approach is very generic and can probably be applied to many design problems outside of automatic passenger counting.

We have also found out that our LSTMs have an upper bound up to which they can count. Our \emph{cumsum layer} thus not only helps create better counts but can be seen as an improvement to LSTMs in general for regression tasks, where calculating the differences in time series (trend elimination) is not applicable. To our knowledge, we are the first to apply this approach to LSTM predictions.

Finally, we hope that our contribution to the NAPC will raise fairness in global revenue sharing and distribution to a new level and that other areas will profit from our results as well.